\documentclass[10pt,twocolumn,letterpaper]{article}

\usepackage[pagenumbers]{cvpr} %

\usepackage{graphicx}
\usepackage{amsmath}
\usepackage{amssymb}
\usepackage{booktabs}

\usepackage{multirow}
\usepackage[normalem]{ulem}
\usepackage{enumitem}
\usepackage{graphicx}
\usepackage{amsmath}
\usepackage{pifont}
\usepackage{dsfont} %
\usepackage[ruled,vlined,linesnumbered]{algorithm2e}
\usepackage{wrapfig}
\usepackage[font=small,labelfont=bf,tableposition=top]{caption}
\usepackage[T1]{fontenc}
\DeclareCaptionLabelFormat{andtable}{#1~#2  \&  \tablename~\thetable}

\def\ryn{\textcolor[rgb]{0.4,0.,1.}}

\usepackage[pagebackref,breaklinks,colorlinks]{hyperref}

\usepackage[capitalize]{cleveref}
\crefname{section}{Sec.}{Secs.}
\Crefname{section}{Section}{Sections}
\Crefname{table}{Table}{Tables}
\crefname{table}{Tab.}{Tabs.}

\def\cvprPaperID{5808} %
\def\confName{CVPR}
\def\confYear{2022}

\begin{document}

\title{Towards Self-Adaptive Pseudo-Label Filtering for Semi-Supervised Learning}

\author{Lei Zhu\\
{\tt\small lzhu68-c@my.cityu.edu.hk}
\and
Zhanghan Ke\\
{\tt\small zhanghake2-c@my.cityu.edu.hk}
\and
Rynson Lau\\
{\tt\small {Rynson.Lau@cityu.edu.hk}}
}
\maketitle

\begin{abstract}

Recent semi-supervised learning (SSL) methods typically include a filtering strategy to improve the quality of pseudo labels.
However, these filtering strategies are usually handcrafted, and do not change as the model is being updated, resulting in a lot of correct pseudo labels being discarded and incorrect pseudo labels being selected during the training process.
In this work, we observe that the distribution gap between the confidence values of correct and incorrect pseudo labels emerges at the very beginning of the training, which can be utilized to filter pseudo labels.
Based on this observation, we propose a Self-Adaptive Pseudo-Label Filter (SPF), which automatically filters noise in pseudo labels 
in accordance with model evolvement
by modeling the confidence distribution throughout the training process.
Specifically, with an online mixture model, we weight each pseudo-labeled sample by the posterior of it being correct, which takes into consideration of the confidence distribution at that time.
Unlike previous handcrafted filters, our SPF evolves together with the deep neural network without manual tuning. 
Extensive experiments demonstrate that incorporating SPF into the existing SSL methods can help improve the performance of SSL, especially when the labeled data is extremely scarce.
\end{abstract}

\section{Introduction}
\label{sec:intro}

\begin{figure*}
\centering
\includegraphics[width=0.9\linewidth]{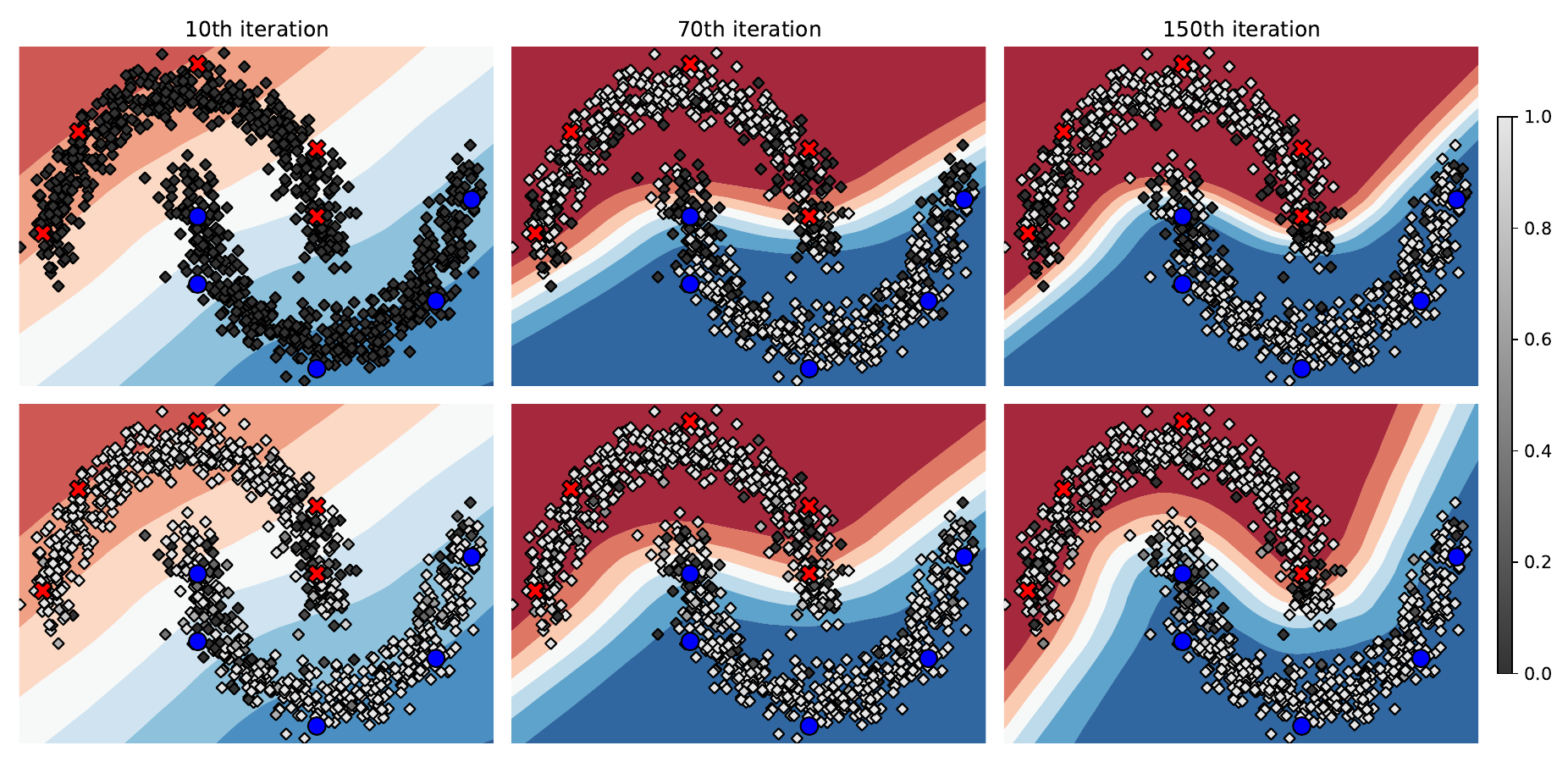}
\caption{Pseudo-label filtering with (top) constant confidence thresholding and (bottom) our proposed Self-Adaptive Pseudo-Label Filter (SPF),
on the ``two moon'' dataset (1000 samples, 5 labeled samples per-class) over the training iterations.
Unlabeled samples are in gray color -- a sample with a darker color indicates that the pseudo label generated for
it has a lower weight.
Using a constant threshold value over the entire training process, almost all unlabeled samples are masked out in the initial stage, causing overfitting on the labeled data. %
Our approach mitigates the problem through self-adaptive pseudo-label filtering, which models the confidence distribution dynamically during the training process.
}
\label{fig:two_moons}
\end{figure*}

Deep neural networks (DNNs) have revolutionized computer vision research and become the de facto framework for many applications, such as image classification~\cite{alexnet, resnet}, object detection~\cite{rcnn, yolo}, and semantic segmentation~\cite{fcn, unet}.
Containing millions of trainable parameters, DNNs are extremely data hungry.
However, collecting manually labeled data requires a tremendous amount of human efforts, which are both tedious and time consuming.
To alleviate such a burden, deep semi-supervised learning (SSL), which leverages a limited amount of labeled data and massive amount of unlabeled data to train DNNs, has attracted more and more attention.

A series of recent state-of-the-art deep SSL methods can be summarized with the teacher-student framework, in which pseudo labels\footnote{In some literature, the term \emph{pseudo label} refers to a constructed ``hard" class assignment for a piece of unlabeled data in pseudo-labeling based methods, excluding the generated ``soft" target in consistency regularization based methods. Here, we use it to refer to both.} are generated with a teacher model when training the student model with unlabeled samples.
However, as the teacher model for producing the pseudo labels is typically constructed from the student model itself (\ie, self-training)~\cite{mean_teacher, fixmatch, hu2021simple, li2020density} or co-trained with the student model (\ie, co-training)~\cite{qiao2018deep_cotraining, dualstudent, nassar2021all, meta_pseudo_labels}, the resulting pseudo labels are not guaranteed to be correct.
Those incorrect pseudo labels may reinforce the incorrectness of the student model and prevent it from learning new knowledge, which is known as \emph{confirmation bias}~\cite{mean_teacher}.

To mitigate the problem of confirmation bias, several handcrafted pseudo-label filtering strategies have been adopted by existing works.
The most related to ours is pseudo-label filtering with confidence thresholding\cite{fixmatch, uda, dash, dualstudent}.
Several works~\cite{fixmatch, uda} have shown that simply keeping the pseudo labels with higher confidence values than a constant threshold (\eg, 0.95) is effective for learning with unlabeled data.
However, we show in this paper a risk induced by using a constant threshold: the DNN tends to overfit the labeled data during the initial stage of the training process, due to the fact that almost all pseudo labels are rejected by the high threshold value. Although more and more unlabeled samples are being accepted gradually as the model evolves, the decision boundary may be more or less already dominated by the labeled data. In other words, the confirmation bias has occurred. As an example, Figure~\ref{fig:two_moons} demonstrates this phenomenon.
A simple improvement is to use a progressive threshold during the training process, as in \cite{dash}. Nonetheless, a handcrafted threshold requires carefully tuning and can be suboptimal.
To handle this problem, we propose a Self-Adaptive Pseudo-Label Filter (SPF), which automatically filter noise in pseudo labels 
in accordance with model evolvement by dynamically modelling the confidence distribution with an unsupervised mixture model throughout the training process.
Unlike~\cite{dash}, our method automatically adapts to the constantly evolving model without manual tuning, and achieves better performances especially in the scenario where the amount of labeled data is extremely scarce.

In summary, the main contributions of our paper include:
\begin{enumerate}[leftmargin=*]
\setlength\itemsep{0em}
    \item We reveal a risk of filtering with a constant confidence threshold: it may cause overfitting of the labeled training data in the early stage, which in turn causes confirmation bias. %
    \item To achieve flexible pseudo-label filtering with model evolvement considered, we propose a self-adaptive approach based on an unsupervised mixture model, which can be plug-and-play into existing SSL methods.
    \item Experimental results on popular SSL benchmarks show that our approach helps improve the performances of existing methods significantly.
\end{enumerate}

\section{Background: Pseudo-Label Filtering in SSL}

\begin{figure}
\centering
\includegraphics[width=\linewidth]{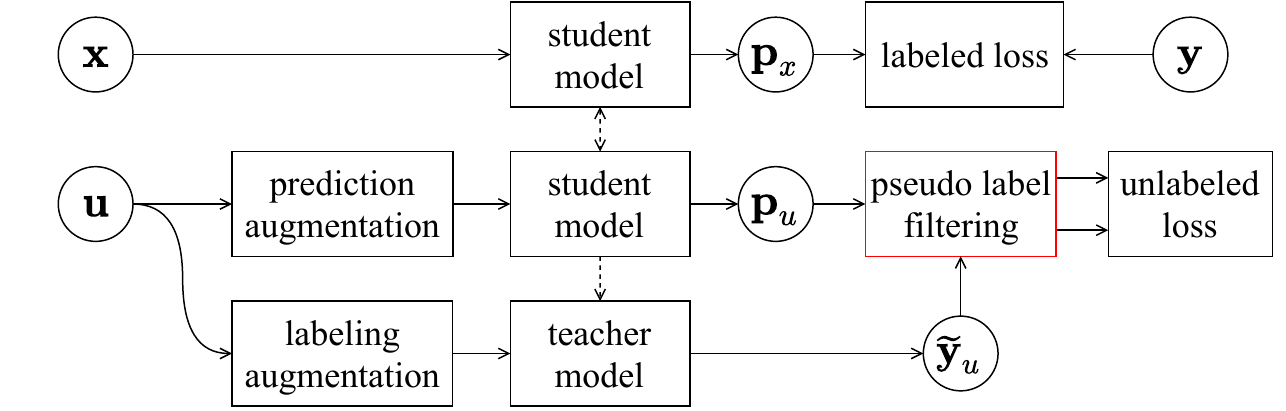}
\caption{
Summarizing the SSL framework. $x$ and $u$ denote labeled and unlabeled data, respectively. The focus of our work is on pseudo-label filtering.
}
  \label{fig:framework_relworks}
\end{figure}

We focus on semi-supervised image classification. A series of recent state-of-the-art methods take advantage of abundant unlabelled data by constructing pseudo labels for them with a teacher model, as shown in Figure~\ref{fig:framework_relworks}.
However, since the teacher model is either constructed from the student model itself or co-trained with it, the resulting pseudo labels are noisy.
To suppress the noise, some simple label filtering strategies have been incorporated as a component.

Formally, given a labeled dataset $\mathcal{X}=\{(x, y)\}$, which contains (image, label) tuples, and an unlabeled dataset $\mathcal{U}=\{u\}$, which contains only images,
a $C$-class image classifier $h_\theta(\cdot)$ is learnt by minimizing the cost function:
\begin{equation}
\label{eq:ssl_prog}
\begin{aligned}
   \mathcal{L}(\theta; \mathcal{X}, \mathcal{U}) =& \sum_{(x,y) \in \mathcal{X}} \ell_S( h_\theta(x), y) + \\
   &\lambda \sum_{u \in \mathcal{U}} w(\tilde{y}_u) \ell_U(h_\theta(\mathcal{A}(u)), \tilde{y}_u),
\end{aligned}
\end{equation}
where $\ell_S(\cdot)$ and $\ell_U(\cdot)$ denote per-sample supervised and unsupervised losses, respectively.
$\lambda \in \mathbb{R}_{+}$ denotes the weight of the unsupervised loss. $\mathcal{A}(\cdot)$ is the data augmentation (prediction augmentation in Figure~\ref{fig:framework_relworks}) applied to the unlabeled data. $w(\tilde{y}_u) \in [0, 1]$ is the filtering weight to suppress the noise in the pseudo labels.
Basically, we expect $w(\cdot)$ to output large values (close to 1) for those correct pseudo labels, but very small values (near 0) for those incorrect ones.

As pseudo labels produced during the initial stage of the training process are usually of low quality, earlier works~\cite{mean_teacher, temporal_ens, mixmatch} typically ramp up the weight of the unlabeled loss (\ie, $\lambda$ in Eq.~\ref{eq:ssl_prog}) from 0 to a pre-defined value.
Such a strategy can be regarded as keeping $\lambda$ constant while using the temporal filtering function:
\begin{equation}
\label{eq:ramp_up}
    w(\tilde{y}_u) = g(t), \ \ \ \forall \tilde{y}_u,
\end{equation}
where $g(t)$ is a monotonically increasing function (\eg, linear\cite{mixmatch} or sigmoid\cite{mean_teacher}) with respect to the training epoch $t$.
However, this strategy is not efficient since all pseudo labels share the same $w(\tilde{y}_u)$ over the same period of time, without considering the correctness of individual samples.

Alternatively, some recent state-of-the-art methods~\cite{uda, fixmatch} adopt a fixed high confidence threshold $\tau$ to filter unreliable pseudo labels.
They keep only those pseudo labels with a confidence score (\ie, the predicted probability for the most possible class)  higher than $\tau$ for network update. Formally, the filtering function that they use is:
\begin{equation}
\label{eq:conf_th}
    w(\tilde{y}_u) = \mathds{1}(\max(\tilde{y}_u) \geq \tau).
\end{equation}
Here, $\mathds{1}(\cdot)$ denotes the indicator function that outputs 1 if the condition is true and 0 otherwise.
In other words, Eq.~\ref{eq:conf_th} estimates a binary weight for each sample, \ie, it considers the per-sample correctness.
Besides, it provides a similar effect as ramping up $\lambda$ in Eq.~\ref{eq:ssl_prog} because there should be more and more pseudo labels with confidence scores exceeding $\tau$ as training progresses~\cite{fixmatch}.

\begin{figure}
\centering
\includegraphics[width=0.9\linewidth]{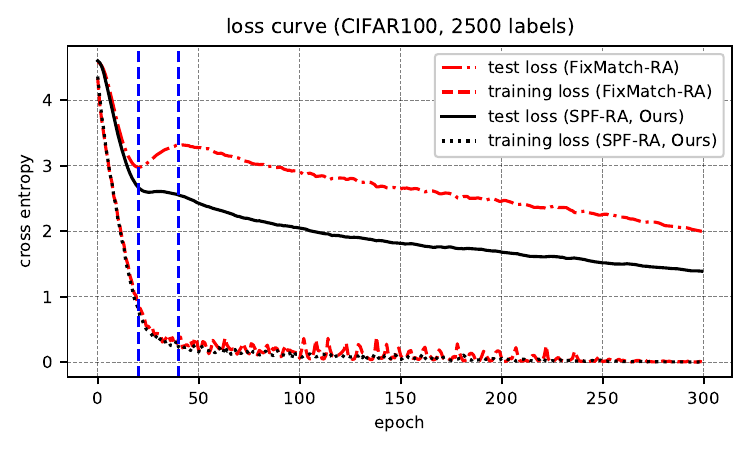}
\caption{ A practical example demonstrating early overfitting on labelled data caused by constant threshold. Note that, FixMatch-RA\cite{fixmatch} uses a constant threshold (0.95) for pseudo-label filtering, and the overfitting occurs between the two vertical dashed lines.
}
  \label{fig:loss_overfitting}
\end{figure}

Nonetheless, using a constant threshold $\tau$ is still problematic as it is not adaptive to model evolvement.
In fact, at the beginning of training, $\tau$ can be too radical that only few unlabeled data contribute to the training. This often makes the model quickly overfit the labeled data, especially when the size of the labeled data is very small.
Besides the example (on the toy dataset) shown in Figure~\ref{fig:two_moons}, we have also observed this phenomenon in practice, as shown in Figure~\ref{fig:loss_overfitting}. In this example, we train WideResNet-28-8~\cite{wrn} 300 epochs on a 2500-label split of CIFAR-100. We monitor the loss (\ie, cross entropy) on the test data and labelled training data. Between $\sim20^{th}$ epoch and $\sim40^{th}$ epoch (the two vertical dashed lines in Figure~\ref{fig:loss_overfitting}), pseudo-label filtering with a constant threshold causes overfitting on the labelled data, as the test loss is increasing while the training loss is decreasing.

There are at least two approaches that may help mitigate this issue.
The first one is to use a progressive threshold during the training process, based on a predefined curriculum~\cite{dash}. However, exploring a suitable tuning curriculum for the threshold is challenging as its pace should be interactively changing as the model evolves.
The second one is to regard the filtering weights $w(\tilde{y}_u)$ as hyper-parameters and optimize them periodically with bi-level optimization~\cite{rys}. However, in this way, a held-out \emph{labeled} validation set is required, which is expensive in the scenario of semi-supervised learning.
Our goal in this paper is to achieve self-adaptive pseudo-label filtering, which takes model evolvement into consideration but without manual tuning and extra data annotations.

\section{Self-Adaptive Pseudo-Label Filtering (SPF)}

\begin{figure}[b]
\centering
\includegraphics[width=\linewidth]{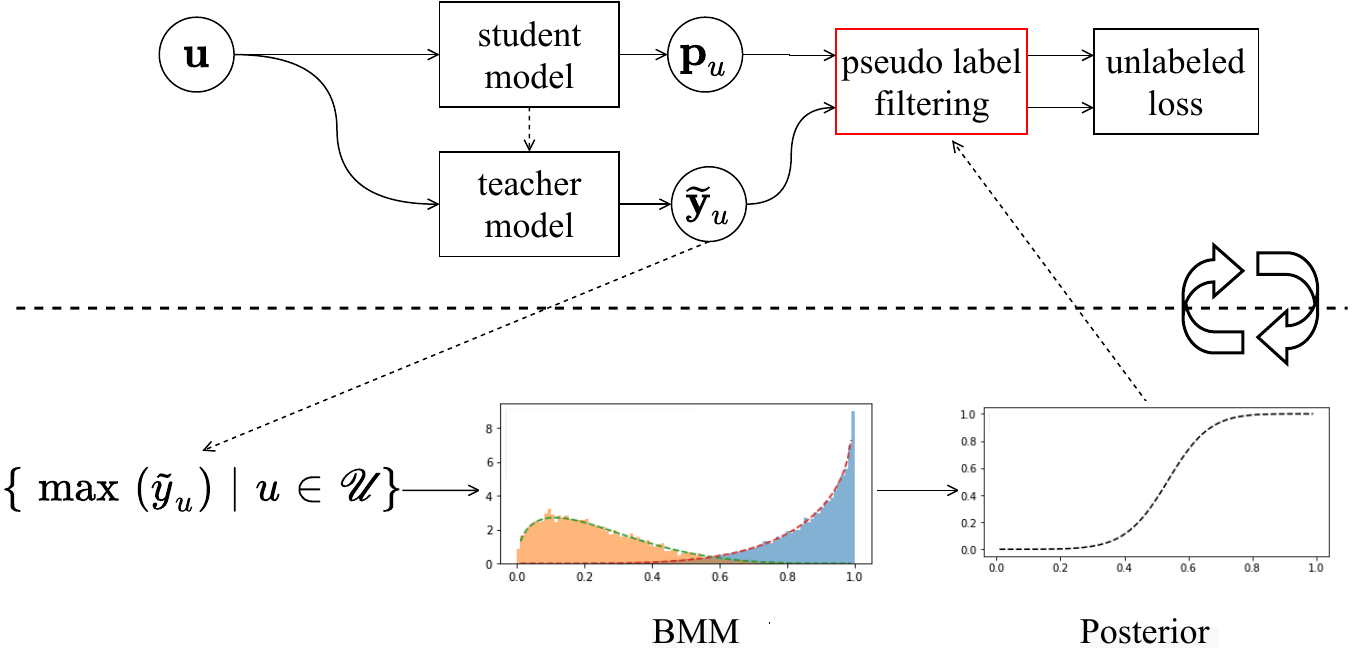}
\caption{
Our adaptive pseudo-label filtering scheme. The DNN and BMM are alternatively updated. For simplicity, some other components in SSL, such as labelled data and data augmentation, are omitted in this diagram.
}
\label{fig:em_upadte}
\end{figure}

Instead of using a constant threshold, our key idea of this work is to filter pseudo labels at each epoch by considering the latest confidence distribution. Specifically, at the end of each epoch, we update a Beta Mixture Model (BMM) with the confidence scores collected during the epoch. In the following epoch, with the updated BMM from the last epoch, we then weight each pseudo-labelled sample with the \emph{posterior} of it being correct when computing the loss. Hence, the deep classifier and the BMM are alternatively updated and evolve together. Figure~\ref{fig:em_upadte} summarizes our method.

\begin{figure*}
\setlength{\tabcolsep}{0.05pt}
\centering
    \begin{tabular}{cccc}
        \includegraphics[width=0.25\linewidth]{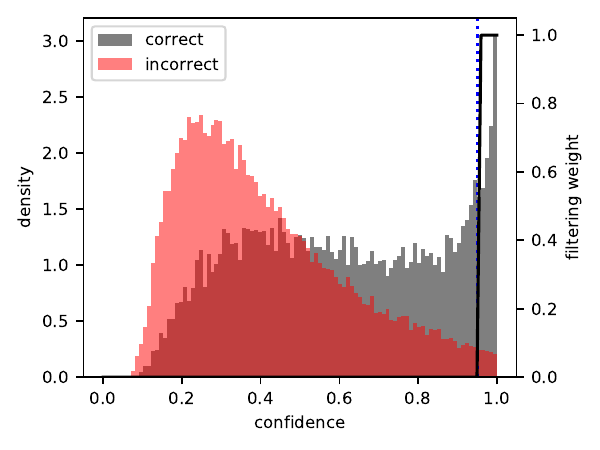}&
		\includegraphics[width=0.25\linewidth]{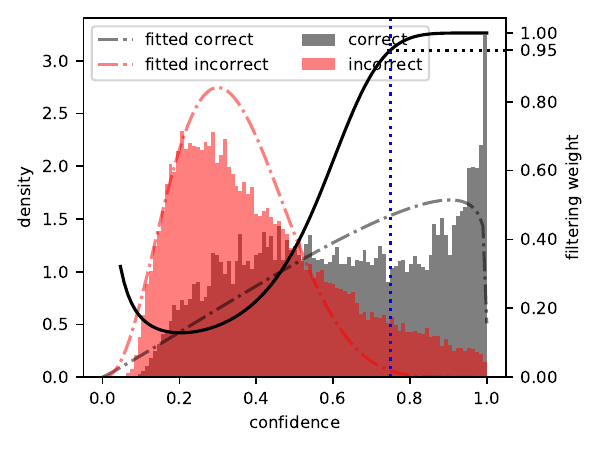}&
		\includegraphics[width=0.25\linewidth]{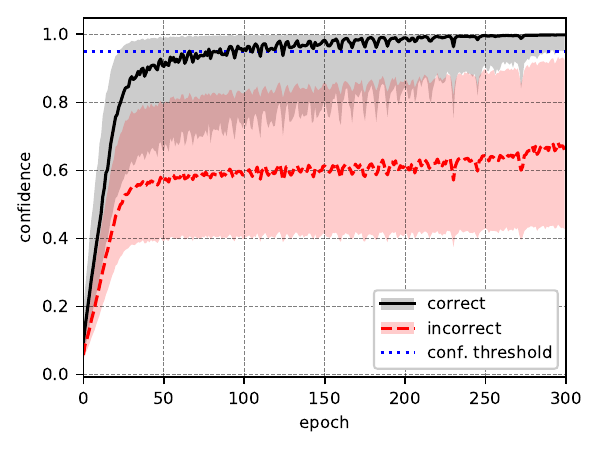}&
		\includegraphics[width=0.25\linewidth]{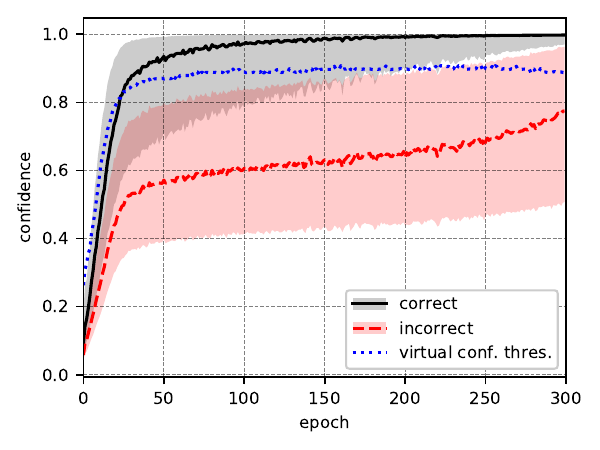} \\
		(a) & (b) & (c) & (d)\\
    \end{tabular}
\caption{Comparison between pseudo-label filtering based on constant thresholding and our probabilistic meta model.
\textbf{(a)} A snapshot of confidence distribution (15$^{th}$ epoch) during training when using constant confidence thresholding (threshold = 0.95). The solid line indicates the filtering weight for pseudo labels.
\textbf{(b)} Similar to (a), but filtering is based on our meta model instead of constant thresholding. The dotted vertical line indicates the virtual confidence threshold (weight $> 0.95$ on the right of it).
\textbf{(c)} Confidence distribution evolvement \wrt training epoch. The solid line and dashed line denote the median of confidence for the correct and incorrect pseudo labels, respectively, while the shaded area denotes the corresponding quartile range. Note that most pseudo labels are masked out at the beginning.
\textbf{(d)} Similar to (c), but filtering is based on our meta model. The virtual confidence threshold shows that our filtering is self-adaptive.
}
\label{fig:early_overfit}
\end{figure*}

\subsection{Modeling Confidence Distribution with BMM}

The mixture model is a widely used \emph{unsupervised} modeling technique~\cite{arazo2019unsupervised,permuter2006study, li2020dividemix} to reveal the statistics of the latent subpopulations (\eg, the correct and incorrect pseudo labels referred to in this paper) within an overall population.
A general mixture model parameterizes the overall distribution as a weighted summation of several homogeneous or inhomogeneous components, each is individually a distribution (\eg, Gaussian and Poisson). Hence, the probability density function (PDF) of a mixture model is written as:
\begin{equation}
\label{eq:mixture_model}
    p(z | \phi ) = \sum_{c=1}^m \gamma_c p(z | c),
\end{equation}
where $m$ is the number of components of the mixture. $p(z | c)$ and $\gamma_c$ are the PDF and weight for each component \ryn{$c$,} respectively. We use $\phi$ to denote the collection of parameters of the mixture model, which includes the weights $\{\gamma_c\}_{c=1}^m$ and parameters of components (\eg, mean and standard deviation if a component is a Gaussian).

In our case, the number of components (\ie, $m$) is determined as 2 since we want to separate pseudo labels into exactly two groups: the correct and incorrect ones.
For each component, we assume that it is a Beta distribution (thus forming a Beta Mixture Model):
\begin{equation}
    p(z | c) = \mathcal{B}(z | \alpha_c, \beta_c) = \frac{\Gamma(\alpha_c + \beta_c)}{\Gamma(\alpha_c)\Gamma(\beta_c)} z^{\alpha_c - 1} (1-z)^{\beta_c - 1},
\end{equation}
where $c=\{1, 2\}$ to index the component. $\alpha_c$ and $\beta_c$ are the two shape parameters controlling the shape of the Beta distribution. $\Gamma(\cdot)$ is the Gamma function.
Note that, as the Beta distribution is defined over the interval $(0, 1)$, it is a natural choice for modelling distribution of probabilities or proportions~\cite{casella2021statistical}, also the confidence scores in our case. In addition, it has a flexible shape that may model both symmetric and skewed distributions~\cite{arazo2019unsupervised}, both of which occur in our case, as shown in Figure~\ref{fig:early_overfit}. In contrast, the popular Gaussian distribution is defined over an unbounded range, and can only be used to model symmetric distributions.

The mixture model is suitable as the meta model for identifying correct and incorrect pseudo labels for three main reasons.
First, it requires no supervision signals (\ie, whether a pseudo label is correct or not), which are not available in practice.
Second, it is simple enough and introduces no extra hyper-parameters and hence requires no manual tuning.
Third, inference and fitting are both efficient, thus causing neglectable overheads.

\begin{algorithm*}
\SetAlgoLined
 \SetKwInOut{KwIn}{Input}
 \KwIn{Neural network $h_\theta(\cdot)$, pseudo-label construction strategy $\mathcal{T}$, labeled training set $\mathcal{X}$, unlabeled training set $\mathcal{U}$, batch size of labeled data $B$, relative batch size of unlabeled data $\mu$, learning rate $\eta$, training epoch $N$}
 \KwOut{The trained neural network $h_\theta(\cdot)$}
 \For{$t=1$ \KwTo $N$}{
    Empty the set of collected confidence scores $\mathcal{S} \leftarrow \emptyset $\;
    \For{$i=1$ \KwTo $\frac{\mid \mathcal{U} \cup  \mathcal{X} \mid}{ (1+\mu) B}$}{
      
     Sample batches $\mathcal{X}^{\prime}=\{(x_b, y_b)\}_{b=1}^{B} \subseteq \mathcal{X}$, $\mathcal{U}^{\prime}=\{u_b\}_{b=1}^{\mu B} \subseteq \mathcal{U}$\;
      
      Construct pseudo labels $\mathcal{Q} \leftarrow \{ \tilde{y_u} \mid u \in \mathcal{U}^{\prime}\} $ with srategy $\mathcal{T}$ \;
     
      Compute per-sample filtering weight according to Eq.~\ref{eq:posterior}\;
      
    Update neural network by gradient descent $\theta \leftarrow \theta - \eta \cdot \nabla_\theta \mathcal{L}(\theta, \mathcal{X}^{\prime}, \mathcal{U}^{\prime})$\;
      
    Collect confidence score  $\mathcal{S} \leftarrow \mathcal{S} \cup \{\max(\tilde{y_u}) \mid \tilde{y_u} \in \mathcal{Q} \}$\;
    }
    Update the BMM with collected statistics $\mathcal{S}$ by Expectation-Maximization\;
 }
 
 \SetKwInOut{KwOut}{Output}
 \caption{SSL with self-adaptive pseudo-label filtering.}
 \label{alg:adaptive}
\end{algorithm*}

\subsection{Alternative Update of DNN and BMM}

With the confidence distribution paramterized by BMM, we are able to compute the \emph{posterior} of a pseudo label being correct. We use the posterior as filtering  weight to suppress the noise in pseudo labels when updating the deep neural network.  As the DNN is continuously evolving, the BMM should be periodically updated to capture the latest confidence distribution. Therefore, the DNN and BMM are alternatively updated as below.

\vspace{1ex}
\noindent
\textbf{Update DNN.} Within the $t^{th}$ epoch, for each sample $u$ with pseudo label $\tilde{y}_u$, we compute the filtering weight $w(\tilde{y}_u)$ for it as the posterior of its pseudo label being correct based on the BMM:
\begin{equation}
\label{eq:posterior}
    \begin{aligned}
    w(\tilde{y}_u) &= p(j | \tilde{y}_u) = 
    \frac{\gamma_j^{(t)} p(z = \max(\tilde{y}_u) | j)}{\sum_{c=1}^2 \gamma_c^{(t)} p(z = \max(\tilde{y}_u) | c)} \\
     s.t. \quad j &= \arg\max_{c=1, 2} \mathbb{E}[p(z| c)] = \arg\max_{c=1, 2} \frac{\alpha_c^{(t)}}{\alpha_c^{(t)} + \beta_c^{(t)}}.
    \end{aligned}
\end{equation}
Note that, the second row which takes expectation on the distribution on $p(z| c)$ in Eq.~\ref{eq:posterior} is to choose the component for the correct pseudo labels, which should have a higher average confidence score. By incorporating Eq.~\ref{eq:posterior} into Eq.~\ref{eq:ssl_prog}, we can then update the DNN parameters $\theta$ with gradient descent as usual.

\vspace{1ex}
\noindent
\textbf{Update BMM.} During the $t^{th}$ training epoch, we collect the confidence scores of the pseudo labels, which are denoted as $S = \{\max(\tilde{y}_u) \mid \ u \in \mathcal{U}\}$. At the end of epoch $t$, we compute new BMM parameters $\phi^{(t+1)} = \{\alpha_1^{(t+1)}, \beta_2^{(t+1)}, \alpha_1^{(t+1)}, \beta_2^{(t+1)}, \gamma_1^{(t+1)}, \gamma_2^{(t+1)}\}$ by fitting $S$ via Expectation-Maximization (see the Supplemental for details). Algorithm~\ref{alg:adaptive} summarizes the steps.

\begin{table*}
  \centering
  \begin{small}
  \resizebox{0.95\textwidth}{!}{
      \begin{tabular}{lcccccc}
        \toprule
                       & \multicolumn{3}{c}{CIFAR-10}   & \multicolumn{3}{c}{CIFAR-100} \\
        \cmidrule(lr){1-1}   \cmidrule(lr){2-4} \cmidrule(lr){5-7} 
        Method         & 40 labels & 250 labels & 4000 labels & 400 labels & 2500 labels & 10000 labels \\
        \cmidrule(lr){1-1}   \cmidrule(lr){2-4} \cmidrule(lr){5-7} 
        $\Pi$-Model & - &  54.26 $\pm$ 3.97 & 14.01 $\pm$  0.38 & - &  57.25 $\pm$ 0.48 & 37.88 $\pm$ 0.11 \\ 
        Pseudo-Labeling & - & 49.78 $\pm$ 0.43 & 16.09 $\pm$ 0.28 & - & 57.38 $\pm$ 0.46 & 36.21 $\pm$ 0.19 \\
        MeanTeacher & - &  32.32 $\pm$ 2.30 & 9.19 $\pm$ 0.19 & - & 53.91 $\pm$ 0.57 & 35.83 $\pm$ 0.24 \\
        MixMatch & 47.54 $\pm$ 11.50 & 11.05 $\pm$ 0.86 & 6.42 $\pm$ 0.10 & 67.61 $\pm$ 1.32 & 39.94 $\pm$ 0.37 & 28.31 $\pm$ 0.33 \\
        UDA & 29.05 $\pm$ 5.93 & 8.82 $\pm$ 1.08 & 4.88 $\pm$ 0.18 & 59.28 $\pm$ 0.88 & 33.13 $\pm$ 0.22 & 24.50 $\pm$ 0.25\\
        ReMixMatch& 19.10 $\pm$ 9.64 & 5.44 $\pm$ 0.05 & 4.72 $\pm$ 0.13 & 44.28 $\pm$ 0.26 & 27.43 $\pm$ 0.31 & 23.03 $\pm$ 0.56\\
        SimPLE & - & - & 5.05 $\pm$ -  & - & - & 21.89 $\pm$ - \\
        \midrule
        FixMatch-RA & 13.81 $\pm$ 3.37 & 5.07 $\pm$ 0.65 & 4.26 $\pm$ 0.05 & 48.85 $\pm$ 1.75 & 28.29 $\pm$ 0.11 & 22.60 $\pm$ 0.12 \\
        RYS  & - & 5.05 $\pm$ 0.12 & 4.35 $\pm$ 0.06 & - & - & - \\
        Dash-RA & 13.22 $\pm$ 3.75 & \textbf{4.56 $\pm$ 0.13} & \textbf{4.08 $\pm$ 0.06} &  44.76 $\pm$ 0.96 & 27.18 $\pm$ 0.21 & 21.97 $\pm$ 0.14 \\
        SPF-RA (Ours) & \textbf{7.46 $\pm$ 2.43} & 5.01 $\pm$ 0.11 & 4.18 $\pm$ 0.07 & \textbf{42.22 $\pm$ 1.13} & \textbf{26.61 $\pm$ 0.19} & \textbf{21.73 $\pm$ 0.14} \\
        \bottomrule
      \end{tabular}
      }
  \end{small}
  \caption{Comparison of error rates on CIFAR-10 with WRN-28-2, and on CIFAR-100 with WRN-28-8.
}
  \label{tab:benchmark}
\vspace{-3mm}
\end{table*}

\begin{table}
  \centering
      \begin{tabular}{l|c|cc}
        \toprule
                                  & \#labels     & MeanTeacher & SPF-MT (Ours) \\
        \midrule
        \multirow{2}{*}{CIFAR-10} & 250  & 32.32 $\pm$ 2.30 & \textbf{29.58 $\pm$ 2.73} \\
                                  & 4000 & 9.19 $\pm$ 0.19 & \textbf{4.54 $\pm$ 0.28} \\
        \midrule
        \multirow{2}{*}{CIFAR-100} & 2500 & 53.91 $\pm$ 0.57 & \textbf{41.54 $\pm$ 0.65} \\ 
                                   & 10000 & 35.83 $\pm$ 0.24 & \textbf{27.03 $\pm$ 0.28} \\
        \bottomrule
      \end{tabular}
  \caption{Combining SPF with MeanTeacher.}
  \label{tab:benchmark_MT}
\vspace{-4mm}
\end{table}

\begin{table}
  \centering

      \begin{tabular}{lcc}
        \toprule
                       & \multicolumn{2}{c}{MiniImageNet} \\
        \cmidrule(lr){1-1}   \cmidrule(lr){2-3}
        Method         & 4000 labels & 10000 labels \\
         \cmidrule(lr){1-1}   \cmidrule(lr){2-3}
         MeanTeacher & 72.51 ± 0.22 & 57.55 ± 1.11 \\
         LP & 70.29 ± 0.81 & 57.58 ± 1.47 \\
         DAG & 55.97 ± 0.62 & 47.28 ± 0.20 \\
         \midrule
         FixMatch-RA & 59.97 $\pm$ 0.10 & 49.13 $\pm$ 0.30\\
         SPF-RA (Ours) & \textbf{50.11 $\pm$ 0.35} & \textbf{43.72 $\pm$ 0.32} \\
        \bottomrule
      \end{tabular}

  \caption{Comparison of error rates on Mini-ImageNet under 4000, 10000 labels with ResNet-18.}
  \label{tab:miniimagenet}
\vspace{-4mm}
\end{table}

\section{Experiments}
\label{sec:exps}

In this section, we first conduct experiments on four image classification datasets (CIFAR-10, CIFAR-100~\cite{krizhevsky2009learning_cifar}, Mini-ImageNet~\cite{NIPS2016_miniimagenet}, and Oxford Flower~\cite{oxford_flower}) to evaluate the effectiveness of our method. We then ablate different components of it to justify our design choices and to demonstrate its effectiveness. Unless stated otherwise, we implement our method SPF-RA by incorporating SPF to FixMatch-RA\cite{fixmatch}.
We report the mean and standard deviation of error rates over 3 runs.
Refer to the Supplemental for the implementation details.

\subsection{Comparison on CIFAR-10 and CIFAR-100}
\label{sec:cifar_cmp}

CIFAR-10~\cite{krizhevsky2009learning_cifar} is a dataset with 60K $32 \times 32$ images evenly distributed across 10 classes. The training set and test set contain 50K and 10K images, respectively. CIFAR-100 is similar to CIFAR-10, but with 100 classes.
On these two datasets, we compare our method SPF-RA with several state-of-the-art methods closely related to ours, including FixMatch-RA~\cite{fixmatch}, Dash-RA~\cite{dash}, RYS\footnote{The authors did not name their algorithm in \cite{rys}. Following \cite{dash}, we name it RYS.}~\cite{rys}. 
Besides, some other popular methods such as $\Pi$-Model~\cite{temporal_ens},  Pseudo-Labeling~\cite{pseudo_label}, MixMatch~\cite{mixmatch}, UDA~\cite{uda}, ReMixMatch~\cite{remixmatch} and SimPLE\cite{hu2021simple} are also included in the comparison.
Following these works, we use Wide-ResNet-28-2~\cite{wrn} for CIFAR-10, and Wide-ResNet-28-8 for CIFAR-100.

We perform our experiments by varying the amount of labeled data, following standard SSL evaluation protocols~\cite{protocol, mixmatch, remixmatch, fixmatch}. Table~\ref{tab:benchmark} shows the performance comparison on top-1 error rate (lower is better). We can see that SPF-RA is the best-performing one in most splits,  which demonstrates the effectiveness of our self-adaptive label filtering approach.
In addition, we observe that the improvement is especially remarkable when the amount of labeled data is extremely scarce. Specifically, when there are only 4 labels per-class, SPF-RA decreases the error rate by more than 6\% on both CIFAR-10 and CIFAR-100, as it can mitigate overfitting on labeled data, which is typically more severe under such an extreme scenario.

As SPF is a plug-and-play component, we also incorporate it into another method, \ie, MeanTeacher, referred to as SPF-MT. The performance comparison of MeanTeacher and SPF-MT on CIFAR-10/100 is listed in Table~\ref{tab:benchmark_MT}. SPF shows clear advantages here again.

\subsection{Comparison on Mini-ImageNet and Flower}
To demonstrate the advantages of SPF on more complex datasets, we conduct experiments on both  Mini-ImageNet~\cite{NIPS2016_miniimagenet} and Oxford Flower Dataset~\cite{oxford_flower}. In these experiments, we use ResNet-18~\cite{resnet} architecture.

Compared to CIFAR-10/100, Mini-ImageNet is a more complex data set as its categories and images are directly sampled from the large-scale ImageNet~\cite{deng2009imagenet}. It consists of 100 classes with 600 images per class. Image resolution is downscaled to $84 \times 84$.
We compare our SPF-RA method with several latest methods, including MeanTeacher~\cite{mean_teacher}, LP~\cite{LP_SSL}, DAG~\cite{li2020density}, and FixMatch-RA~\cite{fixmatch}, under 4k and 10k labels. We show all results in Table.~\ref{tab:miniimagenet}.
SPF-RA achieves notable improvement over the FixMatch-RA baseline and the previous state-of-the-art (DAG). Specifically, SPF-RA reduces the error rate by $9.86\%$ compared with FixMatch-RA and by $5.86\%$ compared with DAG under 4k labels.

Oxford Flower~\cite{oxford_flower} is a fine-grained image classification dataset that consists of 102 different categories of flowers. Each category consists of between 40 and 258 images. Since  fine-grained classification is extremely challenging when the number of labels is scarce, previous works usually did not experiment with such a dataset.
We compare SPF-RA with the supervised only baseline and FixMatch-RA.
As shown in Table.~\ref{tab:flower}, SPF-RA still has a clear improvement on this dataset.
Under the SSL setting with 204 labels, \ie, only two labeled samples per class, SPF-RA outperforms FixMatch-RA by $30.08\%$ on accuracy.

\begin{table}
  \centering
      \begin{tabular}{lcc}
        \toprule
                       & \multicolumn{2}{c}{Oxford 102 Flower} \\
        \cmidrule(lr){1-1}   \cmidrule(lr){2-3}
        Method         & 204 labels & 1020 labels \\
         \cmidrule(lr){1-1}   \cmidrule(lr){2-3}
         SupOnly & 79.69 $\pm$ 0.77 & 40.09 $\pm$ 0.23 \\
         FixMatch-RA & 58.40 $\pm$ 1.24 & 7.24 $\pm$ 0.83 \\
         SPF-RA (Ours) & \textbf{28.32 $\pm$ 2.61} & \textbf{4.96 $\pm$ 0.49} \\
        \bottomrule
      \end{tabular}

  \caption{Comparison of error rates on the Oxford Flower~\cite{oxford_flower} dataset under 204, 1020 labels with ResNet-18.}
  \label{tab:flower}
\vspace{-4mm}
\end{table}

\subsection{Ablation Study}
\label{sec:ablation}

\begin{figure*}[!t]
\begin{minipage}{\textwidth}
\begin{minipage}[b]{0.30\textwidth}
\centering
    \resizebox{0.95\linewidth}{!}{
    \begin{tabular}{lc}
    \toprule
    Method & Top-1 Error$\downarrow$ \\
    \midrule
    CCT &  $60.56 \pm 1.79$\\
    \midrule
    CTR-Linear & $55.23\pm 0.66$\\
    CTR-Sigmoid & $54.45 \pm 1.69$ \\
    CW & $49.36 \pm 1.33$\\
    \midrule
    Meta model (ours) & $\mathbf{47.16 \pm 1.56}$\\
    \bottomrule
    \end{tabular}}
\captionof{table}{Comparison of our meta model with constant confidence thresholding (CCT), confidence weighting (CW) and confidence threshold ramp-up (CTR). Error rates are reported as mean $\pm$ standard deviation over 3 runs on a 400-label split from CIFAR-100.}
\label{tab:abl_conf}
\end{minipage}
\hfill
\begin{minipage}[b]{0.68\textwidth}
    \begin{tabular}{cc}
    \includegraphics[width=0.48\linewidth]{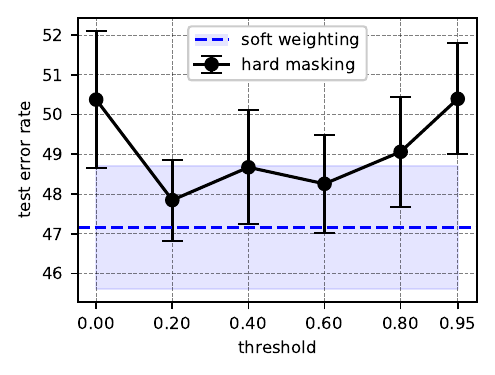} &
    \includegraphics[width=0.48\linewidth]{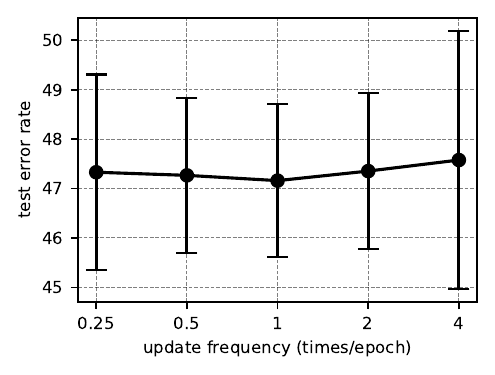}\\
    \end{tabular}
\vspace{-3mm}
\captionof{figure}{Plots of the ablation study on our label filtering approach. \textbf{Left}: Comparison between soft weighting with the posterior of pseudo labels being correct and hard masking with binarized posterior with a varying threshold. \textbf{Right}: Robustness to update frequency of the meta model.}
\label{fig:abl_th_freq}
\end{minipage}
\end{minipage}
\end{figure*}

To further understand how our meta model based label filter helps improve the SSL performances, we perform an extensive ablation study on a moderately challenging setting: CIFAR-100 dataset with 400 labels.  
Due to the number of experiments in our ablation study, we run 800 epochs to save the GPU running time. 
Nevertheless, the model can still converge, as we apply the cosine annealing learning rate decay~\cite{cosine_lr}.
Table~\ref{tab:abl_conf} shows the results. We can see that with our meta model based label filter, SPF-RA achieves a top-1 error rate of $47.16 \pm 1.56 \%$, remarkably outperforming  $60.56 \pm 1.79 \%$ achieved by FixMatch-RA (which uses constant confidence thresholding (CCT)).
In addition, although we reduced the number of training epochs to 800, SPF-RA still outperforms FixMatch-RA, which has more than 10k training epochs and achieves a top-1 error of $48.85 \pm 1.75 \%$ as shown in Table~\ref{tab:benchmark}.
This further demonstrates the superiority of our approach.

\vspace{1ex}
\noindent
\textbf{Confidence weighting (CW) and confidence threshold ramp-up (CTR).} In our filtering approach, we apply a meta model (\ie BMM) to fit the confidence distribution of pseudo labels, and compute the posterior of a pseudo label being correct as its weight based on the meta model.
A naive baseline is directly using the confidence score as the weight without any modelling. We denote this as confidence weighting (CW).
In addition, as shown in Figure~\ref{fig:early_overfit}(d), our method produces the effect of adaptively increasing the threshold. 
Hence, we also compare our method with confidence threshold ramp-up (CTR), which gradually increases the confidence threshold
according to a predefined schedule.
We include two baselines, ``CTR-Sigmoid'' and ``CTR-Linear'', which use the sigmoid and linear ramp-up functions, respectively. Both of them gradually increase the threshold from 0 to 0.95 during the initial 40\% epochs.

We show the results in Table~\ref{tab:abl_conf}.
First, CW achieves a lower error rate than constant confidence thresholding (CCT).
This may be due to the fact that CW on the one hand provides a natural filter that is able to assign higher weights to the correct pseudo labels than to the incorrect ones, and on the other hand avoids overfitting on labeled data by taking every unlabeled data into consideration even in the early stage.
Our meta model based filter further improves the performance by giving even higher weights to those high-confidence pseudo labels and lower weights to those low-confidence ones, by taking consideration of the whole confidence distribution.
Second, both CTR-Sigmoid and CTR-Linear outperform the plain baseline CT, indicating that it is beneficial to consider model evolvement.
Although tuning the ramp-up schedule may help improve the performance, it requires an exhaustive trial-and-error loop.
Instead, ours method, which is based on the meta model, is self-adaptive and performs significantly better.

\vspace{1ex}
\noindent
\textbf{Hard masking and soft weighting.} Just like in confidence thresholding~\cite{fixmatch, dash}, it is also possible to binarize the posterior using some threshold, yielding a ``hard'' masking version of our approach. 
We study the effect of using different thresholds. Figure~\ref{fig:abl_th_freq}~(Left) shows the results.
Using a high threshold value of 0.95 or extremely low threshold value of 0 significantly increases the test error by more than 3.2\%, indicating that the quality and quantity of the pseudo labels are both important.
In addition, although the best quantity-quality trade-off can be achieved by using a threshold value of 0.2, it still cannot outperform ``soft'' weighting.
In summary, hard masking is not able to outperform soft weighting. Instead, it introduces an extra hyper-parameter, \ie, a threshold value.

\vspace{1ex}
\noindent
\textbf{Robustness of the meta model to update frequency.} 
In Algorithm~\ref{alg:adaptive}, we update the meta model once per-epoch.
However, it is possible to change the update frequency.
Note that while a more frequent update allows the meta model to respond more promptly to the learning state of the network, it sacrifices the quantity of the confidence scores for the pseudo labels
collected to fit the meta model, causing inaccurate estimation of the meta model parameters.
Here, we try increasing/decreasing the update frequency by a factor of two and four.
Figure~\ref{fig:abl_th_freq}~(Right) shows the test error w.r.t. the update frequency.
We can see that while the lowest test error is achieved at the default frequency (once per-epoch), the performance degrades only slightly if we either increase or decrease the update frequency of the meta model.
This means that our approach is robust to the update frequency of the meta model.

\begin{figure*}
\centering
\includegraphics[width=0.98\linewidth]{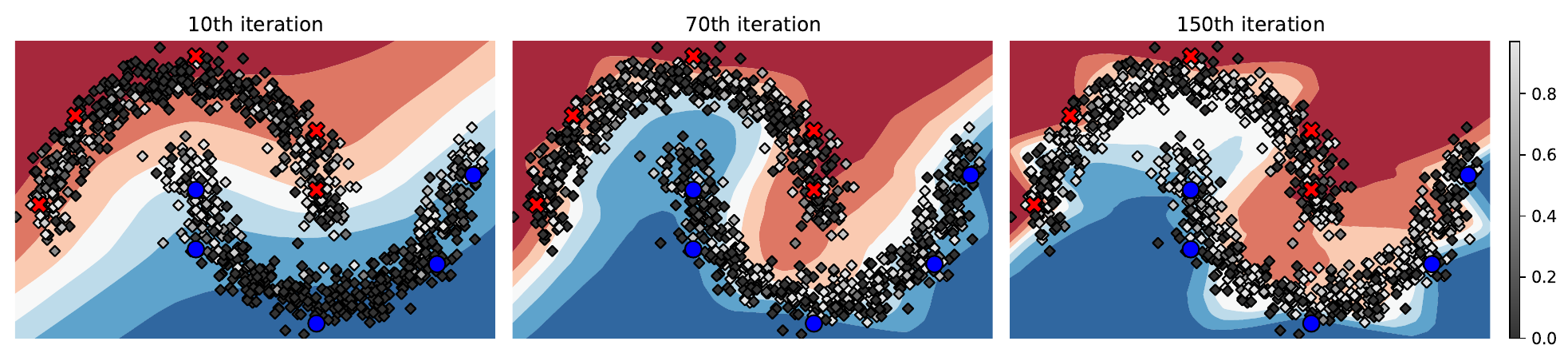}
\caption{
Failure case. We repeat the experiment in Figure~\ref{fig:two_moons} with the same setting, except by increasing the learning rate from $5e-4$ to $5e-3$. Our method will then fail to prevent the model from overfitting as the meta model cannot catch up with the pace of model evolvement, and it finally collapses. Unlabeled samples are in gray color; the intensity of a sample indicates the weight of its pseudo label.
}
\label{fig:two_moons_failure}
 \vspace{-1mm}
\end{figure*}

\section{Related Works}

\noindent
\textbf{Semi-supervised Learning (SSL).} SSL is a transversal task for different domains with a huge diversity of approaches. We focus on methods closely related to ours, \ie, those using deep neural networks for image classification. When learning from labeled data, these methods use a cross-entropy (or similar) loss as in standard supervised learning. 
When learning from unlabeled data, pseudo labels are constructed with a teacher model.
The teacher model can be a fixed pretrained model~\cite{pseudo_label}. However, a fixed teacher is not able to correct itself, leading to inferior performances to the student model.
Hence, recent approaches typically construct the teacher model from the student model itself on the fly~\cite{fixmatch, mixmatch, hu2021simple, li2020density}, or by co-training together with the student model~\cite{qiao2018deep_cotraining, dualstudent}.
Nonetheless, it is still not guaranteed that the generated pseudo labels are correct as the teacher model itself is not perfect.
Those incorrect pseudo labels lead the student model to reinforce its incorrectness, hence preventing it from learning new knowledge, which is known as confirmation bias~\cite{mean_teacher}.

As a pseudo-label filtering method, our approach aims to suppress noise in pseudo labels dynamically as the model evolves, and can easily be incorporated into the existing SSL methods to help improve their performances.
\vspace{1ex}

\noindent
\textbf{Pseudo-label filtering.} To mitigate the confirmation bias, existing SSL methods are typically equipped with some pseudo-label filtering strategy.
A simple and straightforward approach is to gradually ramp up the weight of unlabeled loss\cite{mean_teacher,mixmatch, remixmatch}, which can be viewed as filtering out pseudo labels produced during the initial stage of the training process. These pseudo labels are expected to be of low quality as the teacher model is in its infancy.
However, this approach is not efficient as it treats all pseudo labels produced at a specific period of time equally, without considering per-sample correctness.
Alternatively, some SSL methods~\cite{uda, fixmatch} apply a constant confidence threshold (\eg, 0.95) for pseudo-label filtering.
Only those pseudo labels with a confidence score (\ie, the predicted probability for the most possible class) higher than the threshold is kept for network update.
Such a method is further improved by using a progressive threshold over the training process~\cite{dash}. 
Although progressive thresholding is effective, it raises a question: what is the problem with a constant high threshold while a high threshold seems to reject more incorrect pseudo labels?
Besides, the approach proposed in \cite{dash} still relies on a predefined schedule to adjust the threshold \wrt the training progress. Hence, it requires careful tuning and may result in suboptimal performance of the final model.

In this paper, we reveal a risk caused by a high constant threshold: it may cause the deep network to overfit the labeled data in the initial stage, as it masks out majority of the pseudo-labelled data.
To address this problem, we propose a fully self-adaptive filtering strategy, which is based on an unsupervised mixture model and hence requires no manual tuning.

\section{Conclusion}
\label{sec:conclusion}

In this paper, we have discussed the drawbacks of existing pseudo-label filtering approaches for SSL.
In particular, we have shown that pseudo-label filtering with a constant threshold, which is commonly used in recent state-of-the-art SSL methods, does not take model evolvement into consideration. 
This causes overfitting on the labeled training data in initial training stage.
While progressive thresholding may help improve pseudo-label filtering, a handcrafted schedule requires carefully tuning and may be suboptimal.
To mitigate this problem, we have proposed a self-adaptive label filtering approach by learning a meta model online.
It is interactively changing as the model evolves.
Our method can be plug-and-play into existing SSL methods.
We experimentally show that our proposed method helps improve the performances of existing SSL methods, especially when the labelled data is extremely scarce.

Our meta model does have limitations. As it is not updated in real time, our approach may not be stable if the DNN evolves rapidly (\eg, if a large learning rate is applied). %
Note that our meta model is updated with the statistics collected from the last epoch. The confidence distribution captured from it has a small lag to the actual instant confidence distribution. If the model evolves smoothly, which is typically the case, the lag has very small effects.
However, if the model evolves rapidly, the filtering weight will be wrongly computed with a stale meta model.
Figure~\ref{fig:two_moons_failure} demonstrates such a failure on the toy dataset. In practice, this kind of problem can occur if an extremely large learning rate is used.

\clearpage

\appendix
\section*{\Large{Appendix}}

In this Appendix, we first provide the Expectation-Maximization procedure for Beta Mixture Model (Section~\ref{app:em_bmm}) and implementation details for our experiments (Section~\ref{app:imp_details}).
We then discuss more details about feature and model selection for our meta model (Section~\ref{app:feat_model_sel}) and overheads caused by the meta model (Section~\ref{app:overhead}).

\section{Expectation-Maximization for BMM}
\label{app:em_bmm}

We use an Expectation-Maximization procedure to estimate the parameters of our meta model, \ie the two-component BMM.
Specifically, we introduce a latent variable $W \in \mathbb{R}^{n \times 2}$, in which $W_{ij}$ is the responsibility of component $j$ for the sample point $z_i$ such that $\sum_{j=1}^{2} W_{ij} = 1, \forall i$.
In E-step, we fix the visible parameter collection $\phi = \{ \alpha_1, \beta_1, \alpha_2, \beta_2, \gamma_1, \gamma_2 \}$ of BMM, and compute the latent variable $W$ using Bayes rule:
\begin{equation}
	W_{ij} = p(j | s_i) =  \frac{\gamma_j \mathcal{B}(z_i | \alpha_j, \beta_j)}{\sum_{k=1}^2 \gamma_k \mathcal{B}(z_i| \alpha_k, \beta_k)}, \\
\end{equation}
where $i = 1, 2, ..., n$ and $j=1, 2$ index the rows and columns of the latent variable $W$ respectively.
In M-step, given $W$ computed in the E-step, we first compute the mean $\mu_j$ and variance $\sigma_j^{2}$ for each component:
\begin{equation}
	\mu_j = \frac{\sum_{i=1}^{n} W_{ij} s_i}{\sum_{i=1}^{n} W_{ij}}, \quad
	\sigma_j^{2} = \frac{\sum_{i=1}^{n} W_{ij} (s_i - \mu_j)^2}{\sum_{i=1}^{n} W_{ij}}.
\end{equation}
Then we update the shape parameters of each component, \ie $\alpha_j$ and $\beta_j$, using the method of moments:
\begin{equation}
	\alpha_j = \mu_j(\frac{\mu_j(1-\mu_j)}{\sigma_j^{2}} -1), \quad
    \beta_j =  (1-\mu_j)(\frac{\mu_j(1-\mu_j)}{\sigma_j^{2}} -1).
\end{equation}
Finally, the weights of components $\gamma_j$ are calculated as:
\begin{equation}
	\gamma_j = \frac{1}{n} \sum_{i=1}^{n} W_{ij}.
\end{equation}
In our experiments, following~\cite{arazo2019unsupervised}, we run the above E- and M-steps for 10 iterations in each fitting.

\section{Implementation Details}
\label{app:imp_details}

\begin{figure*}
\setlength{\tabcolsep}{0.1pt}
\centering
    \begin{tabular}{ccc}
        \includegraphics[width=0.33\linewidth]{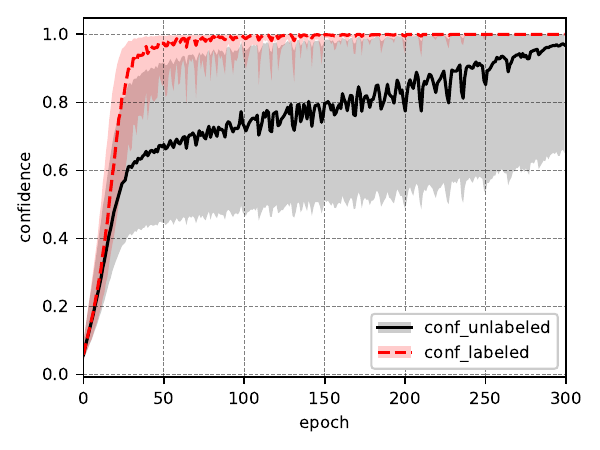}&
		\includegraphics[width=0.33\linewidth]{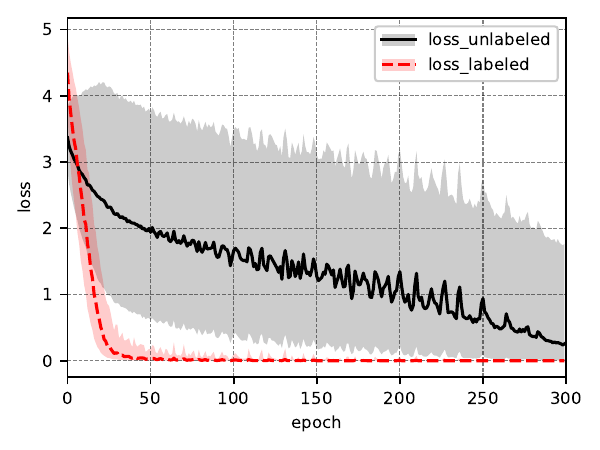}&
		\includegraphics[width=0.33\linewidth]{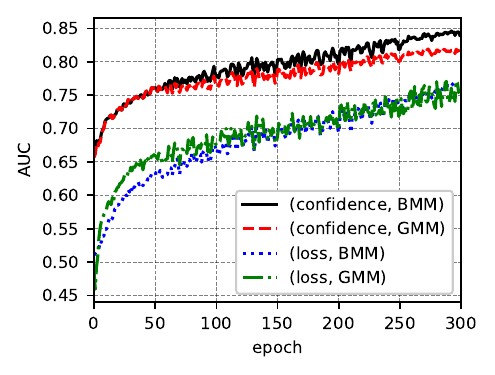} \\
		{(a)} &{(b)} & {(c)} \\
	\end{tabular}
\caption{Feature and model selection for the meta model.
\textbf{(a)} The distributional gap between confidence scores of labeled data and unlabeled data. The lines denote the medians of confidence, while the shaded areas denote quartile ranges.
\textbf{(b)} The distributional gap between loss values of labeled data and unlabeled data.
\textbf{(c)} AUROC evolution \wrt training epoch under different (feature, model) combinations.
}
\label{fig:feat_model_sel}
\end{figure*}

In this section, we provide implementation details for the experiments. All experiments are performed on 4 single-GPU machines, each of which is equipped with an NVIDIA 2080Ti 11GB GPU.

\subsection{CIFAR-10/100 Experiments}

\begin{table}
  \centering
  \begin{small}
  \resizebox{0.98\linewidth}{!}{
      \begin{tabular}{lcccc}
        \toprule
                 & epochs &   400 labels & 2500 labels & 10000 labels \\
        \midrule
        FixMatch-RA & 10k& 48.85 $\pm$ 1.75 & 28.29 $\pm$ 0.11 & 22.60 $\pm$ 0.12 \\
        \midrule
        FixMatch-RA & 2k & 52.40 $\pm$ 1.07 & 29.83 $\pm$ 0.83 & 23.76 $\pm$ 0.20 \\
        SPF-RA & 2k & \textbf{42.22 $\pm$ 1.13} & \textbf{26.61 $\pm$ 0.19} & \textbf{21.73 $\pm$ 0.14} \\
        \bottomrule
      \end{tabular}
      }
  \end{small}
  \caption{Extra experiments on CIFAR-100 to compare SPF-RA with FixMatch-RA. Note that, numbers in the first row are copied from \cite{fixmatch} and they are produced from 10k epochs' training.}
  \label{tab:extra_cifar100}
\end{table}

Following FixMatch~\cite{fixmatch}, we use SGD optimizer with Nesterov momentum to optimize the deep neural network.
Momentum of the optimizer is set to 0.9. 
We apply a cosine learning rate decay schedule as $ \eta = \eta_0 \cos(\frac{7 \pi k}{16 K})$, where $\eta_0$ is the initial learning rate, $k$ is the current training step, and $K$ is the total training step. The initial learning rate $\eta_0$ is set as 0.03.

For CIFAR-10 experiments, we use WideResNet-28-2 architecture, and set the weight decay $5e-4$.
In each training step, we feed the network a mini-batch containing 64 labeled samples and $7 \times 64 = 448$ unlabeled samples. 
The network is trained for 10k epochs.
Such a setup is identical to it used in FixMatch~\cite{fixmatch}.

For CIFAR-100 experiments, we use WideResNet-28-8 architecture and correspondingly change weight decay to $5e-3$.
To fit mini-batches into 11GB GPU memory, we decrease the batch size from 512 to 256.
To save GPU running time, we also decrease the training epoch.
We find that training for 2k epochs is sufficient for our algorithm (SPF-RA) to outperform FixMatch-RA~\cite{fixmatch}.
For fair comparison, we also run FixMatch-RA under the same setting.
The comparison of top-1 error is shown in Table~\ref{tab:extra_cifar100}.
We can see that FixMatch-RA has higher error rates when the training epoch is decreased to 2k. Hence our SPF-RA (trained for 2k epochs) still performs best.

\subsection{Mini-ImageNet and Oxford Flower}

For the models trained on Mini-ImageNet and Oxford Flower, we apply the ResNet-18~\cite{resnet} architecture. We use SGD with a learning rate of 0.01. The batch size is 128, and the weight decay is $1e^{-3}$. The training epoch is 1000 for Mini-ImageNet and 300 for Oxford Flower. The learning rate decay schedule is the same as that used in the CIFAR experiments.

\section{Feature and Model Selection for Meta Model}
\label{app:feat_model_sel}

We fit a probabilistic meta model online to help identify correct and incorrect pseudo labels in our approach.
In this section, we provide more details about feature and model selection for the meta model.

For feature selection, an alternative to the confidence score is loss value, which has been shown helpful for recognizing noisy manual annotations in dataset~\cite{arazo2019unsupervised,li2020dividemix}. 
For model selection, as we use the Beta Mixture Model (BMM) in our approach, a straightforward alternative is the more popular Gaussian Mixture Model (GMM).
We do not consider training a supervised model with meta statistics collected from labeled training data since there is a distributional gap between it of labeled and unlabeled training data as shown in Figure~\ref{fig:feat_model_sel}(a) and Figure~\ref{fig:feat_model_sel}(b).
Such a gap makes the supervised model cannot be well transferred to unlabeled training data.

As the meta model is applied for online binary classification (\ie identifying correct and incorrect pseudo labels), we compare different feature and model combinations in terms of the Area Under the Receiver Operating Characteristic (AUROC) evolution during training.
Note that a higher AUROC score means better classification performance.
Empirically, acceptable discrimination requires an AUROC score higher than 0.7~\cite{hosmer2013applied}.
Figure~\ref{fig:feat_model_sel}(c) shows the results.
First, we can see that the correct and incorrect pseudo labels are more discriminative with confidence score than loss value.
Intuitively, this is because the unsupervised loss is more dominated by the perturbation strength of inputs or the gap between teacher model and student model, instead of the correctness of the pseudo label.
Second, when using confidence as the input feature for our meta model, though the BMM and GMM produce similar classification performance at the beginning of training, BMM gradually outperforms GMM by a considerable margin.
This is due to the confidence distributions become more and more skewed towards 1, and Beta distribution is flexible to handle such skewness.
On the contrary, Gaussian distribution is always in symmetric shape.

\section{Overheads of the Meta Model}
\label{app:overhead}

\begin{table}
  \centering
  \begin{small}
      \begin{tabular}{lcc}
        \toprule
                 & Ave. batch time & Ave. epoch time\\
        \midrule
        FixMatch-RA & 0.61 s & 120.22 s\\
        SPF-RA & 0.62 s & 121.85 s \\
        \midrule
        Overheads & 0.01 s & 1.63 s \\
        \bottomrule
      \end{tabular}
  \end{small}
  \caption{Overheads caused by our meta model. We train WideResNet-28-8 on CIFAR-100 dataset, each mini-batch contains 256 samples (32 labeled and 224 unlabeled). Experiments are performed on a single NVIDIA 2080Ti GPU.}
  \label{tab:overheads}
\vspace{-6mm}
\end{table}

The fitting and inference of our meta model can both cause overheads.
Note that the fitting happens at the end of each epoch, and the inference happens every training step when we feed DNN a mini-batch.
Hence we compare both average epoch time and batch time in Table~\ref{tab:overheads}.
We can see that the overheads caused by our meta model are less than 2\%, which is neglectable.

{\small
\bibliographystyle{ieee_fullname}
\bibliography{egbib}
}

\end{document}